\documentclass{article}

\PassOptionsToPackage{numbers,sort&compress}{natbib}
\usepackage[final]{neurips_mlsb_2024}
\usepackage[utf8]{inputenc} 
\usepackage[T1]{fontenc}    
\usepackage{hyperref}       
\usepackage{url}            
\usepackage{booktabs}       
\usepackage{amsfonts}       
\usepackage{nicefrac}       
\usepackage{microtype}      
\usepackage{xcolor}         
\usepackage{graphicx}

\usepackage{setspace}
\usepackage{algorithm}
\usepackage{algorithmic}
\newcommand\norm[1]{{\left\lVert#1\right\rVert}} 

\usepackage{multirow}
\usepackage{makecell}
\usepackage{amsmath,amssymb}
\DeclareMathAlphabet{\bbold}{U}{bbold}{m}{n}
\newcommand{\id}{\ensuremath{\bbold{1}}}

\definecolor{customgreen}{HTML}{009E73}
\definecolor{customgold}{HTML}{E69F00}

\title{Exploring Discrete Flow Matching for 3D De Novo Molecule Generation}

\author{%
  Ian Dunn \\
  Dept. of Computational \& Systems Biology\\
  University of Pittsburgh\\
  Pittsburgh, PA 15260 \\
  \texttt{ian.dunn@pitt.edu} \\
  \And
  David Ryan Koes \\
  Dept. of Computational \& Systems Biology\\
  University of Pittsburgh\\
  Pittsburgh, PA 15260 \\
  \texttt{dkoes@pitt.edu} \\
}

\begin{document}

\maketitle

\begin{abstract}
    Deep generative models that produce novel molecular structures have the potential to facilitate chemical discovery. Flow matching is a recently proposed generative modeling framework that has achieved impressive performance on a variety of tasks including those on biomolecular structures. The seminal flow matching framework was developed only for continuous data. However, \textit{de novo} molecular design tasks require generating discrete data such as atomic elements or sequences of amino acid residues. Several discrete flow matching methods have been proposed recently to address this gap. In this work we benchmark the performance of existing discrete flow matching methods for 3D \textit{de novo} small molecule generation and provide explanations of their differing behavior. As a result we present FlowMol-CTMC, an open-source model that achieves state of the art performance for 3D \textit{de novo} design with fewer learnable parameters than existing methods. Additionally, we propose the use of metrics that capture molecule quality beyond local chemical valency constraints and towards higher-order structural motifs. These metrics show that even though basic constraints are satisfied, the models tend to produce unusual and potentially problematic functional groups outside of the training data distribution. Code and trained models for reproducing this work are available at \url{https://github.com/dunni3/FlowMol}.
\end{abstract}


\section{Introduction}

Deep generative models that can directly sample molecular structures with desired properties have the potential to accelerate chemical discovery by reducing or eliminating the need to engage in resource-intensive, screening-based  discovery paradigms. Moreover, generative models may improve chemical discovery by enabling multi-objective design of chemical matter. In pursuit of this idea, there has been recent interest in developing generative models for the design of small-molecule therapeutics \cite{huang_dual_2024,guan_3d_2023,schneuing_structure-based_2023,peng_pocket2mol_2022,liu_generating_2022,torge_diffhopp_2023,igashov_equivariant_2024,dunn_accelerating_2023}, proteins \cite{watson_novo_2023,bennett_atomically_2024,ingraham_illuminating_2023}, and materials \cite{zeni_mattergen_2024}. The most popular approach for these tasks has been to apply diffusion models \cite{sohl-dickstein_deep_2015,ho_denoising_2020,song_score-based_2021} to point cloud representations of molecular structures. 

Flow matching \cite{lipman_flow_2023, tong_improving_2023, albergo_stochastic_2023, liu_flow_2022} is a generative modeling framework that generalizes diffusion models while being both simpler and providing more flexibility in model design. This flexibility has enabled flow matching to improve over diffusion in some cases, demonstrating impressive results \cite{ma_sit_2024,bose_se3-stochastic_2024,yim_fast_2023,huguet_sequence-augmented_2024}. However, the seminal flow matching formulation is only designed for continuous data, constraining the domain of problems that can be modeled under this framework. To address this gap, several recent works have proposed discrete flow matching (DFM) methods. Existing discrete flow matching methods fall into two general categories: applying continuous flow matching to a continuous embedding of discrete data \cite{dunn_mixed_2024,davis_fisher_2024,stark_dirichlet_2024,cheng_categorical_2024,eijkelboom_variational_2024} , or defining flows on discrete state spaces with Continuous Time Markov Chains (CTMC) \cite{campbell_generative_2024,gat_discrete_2024}. However, there has yet to be a controlled comparison of these distinct approaches for discrete generative modeling. 

Our contributions are as follows:

\begin{enumerate}
    \item A direct comparison of discrete flow matching methods on \textit{de novo} molecule generation, controlling for model architecture and training procedures.
    \item FlowMol-CTMC, a flow matching model that achieves state of the art molecular validity while using fewer learnable parameters than baseline methods.
    \item Novel metrics that capture molecule quality beyond local chemical valency constraints and towards higher-order structural motifs. 
\end{enumerate}

\begin{figure}[tb]
    \centering
    \label{fig:ga}
    \includegraphics[width=0.7\textwidth]{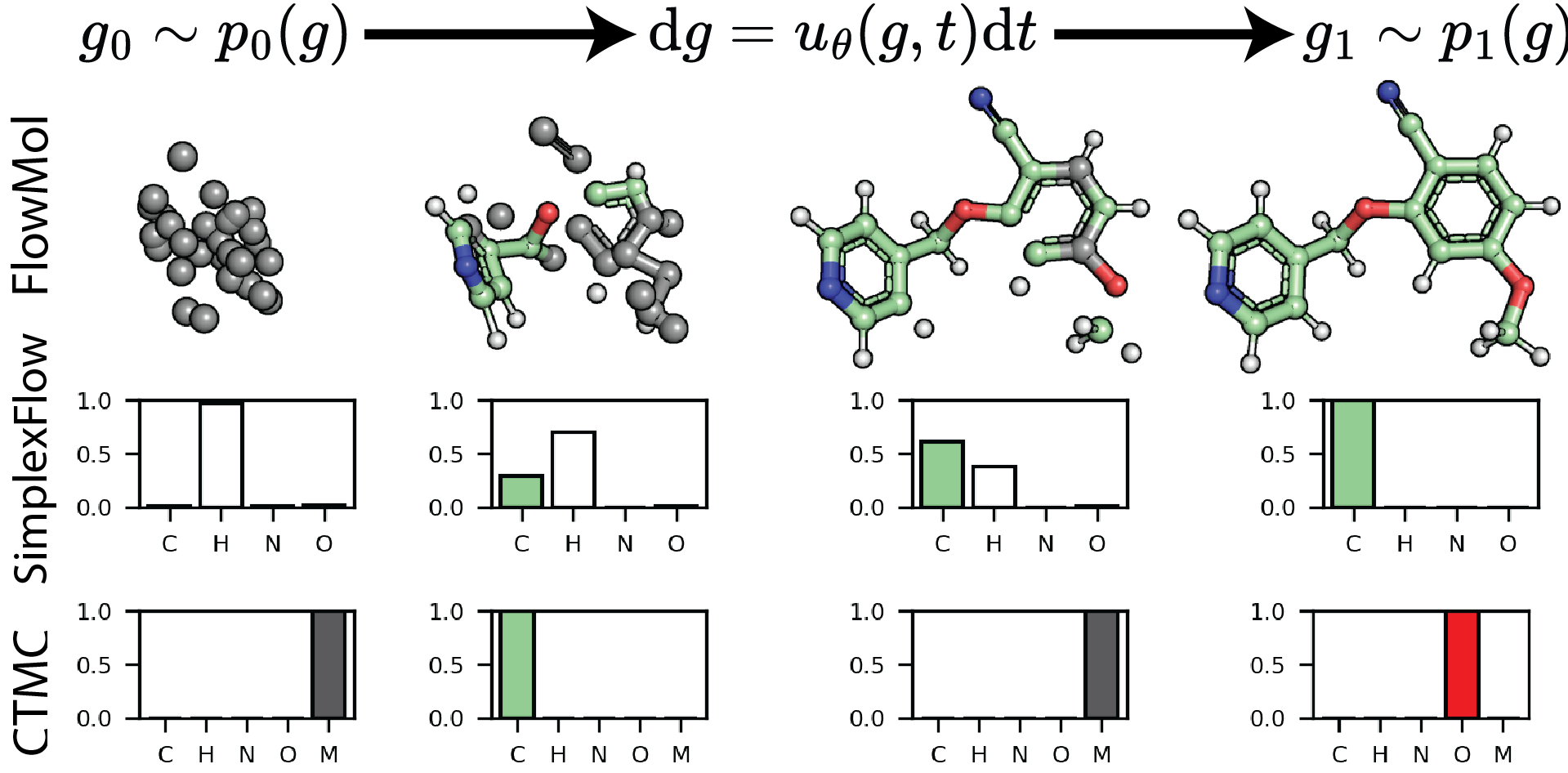}
    \caption{\textbf{Overview} \textit{Top:} We adapt the flow matching framework for unconditional 3D molecule generation and explore the use of different discrete flow matching methods. This CTMC trajectory shows masked atoms in gray. \textit{Middle:} Trajectory of the atom type vector for a single atom under SimplexFlow, a variant of continuous flow matching developed for categorical variables. Atom type flows lie on the probability simplex. \textit{Bottom:} Trajectory of an atom type vector for a CTMC flow. Atom types jump between the mask state and real atom types.}
\end{figure}

\section{Background}

Flow matching \cite{tong_improving_2023,albergo_stochastic_2023,lipman_flow_2023, liu_flow_2022} prescribes a method to interpolate between two distributions $q_{\text{source}}$, $q_{\text{target}}$ by modeling a set of time-dependent marginal distributions $p_t$ having the property that $p_0 = q_{\text{source}}$ and $p_1 = q_{\text{target}}$. The source distribution is
typically a simple prior and the target distribution a complex data distribution that cannot be sampled easily. Given access to a conditioning variable $z \sim p(z)$, the marginal process $p_t$ can be described as an expectation over conditional probability paths:


\begin{equation}
    p_t(x) = \mathbb{E}_{p(z)} \left[  p_t(x|z)  \right]
\end{equation}

The conditioning variable is generally taken to be either the final value $z=x_1$ or pairs initial and final points $z=(x_0,x_1)$. Conditional probability paths are chosen such that they can be sampled in a simulation free manner for any $t \in [0,1]$

\paragraph{Continuous Flow Matching} In the continuous case $p_t(x)$ can be sampled by numerical integration of a learned vector field $u_t(x) = \frac{dx}{dt}$. This vector field can be the direct output of a neural network or it can be parameterized as a function of an optimal denoiser $\hat{x}_1(x_t)$ that minimizes the tractable training objective \eqref{eq:cont-fm-objective}. The latter approach has been found to be more effective for molecular structures \cite{huguet_sequence-augmented_2024,dunn_mixed_2024}. 

\begin{equation} \label{eq:cont-fm-objective}
    \mathcal{L} \propto \mathbb{E}_{t,p(z),p_t(x_t|z)}\left[\norm{\hat{x}_1(x_t) - x_1}\right]
\end{equation}

\paragraph{Continuous Approaches to Discrete Flow Matching}

The simplest approach for flow matching on discrete data is to build continuous flows from a Gaussian prior to one-hot vectors \cite{song_equivariant_2023,eijkelboom_variational_2024}; this approach makes no accommodation for the discrete nature of the data. Alternatively, one can define a continuous embedding of discrete data, and then perform continuous flow matching on the embedding. This is the approach of SimplexFlow \cite{dunn_mixed_2024} and Dirichlet Flows \cite{stark_dirichlet_2024} where the continuous representation of choice is the probability simplex. Additional works have proposed Riemannian flow matching on the probability simplex \cite{davis_fisher_2024,cheng_categorical_2024}; these methods also fall into the category of continuous flow matching on continuous representations of discrete data, although we do not implement these methods in our work. 

\paragraph{CTMC} \citet{campbell_generative_2024} and \citet{gat_discrete_2024} develop a flow matching method for sequences of discrete tokens $x = \{x^i\}_{i=1}^N$ where each sequence element $x^i \in \{1,2,\dots,D\}$ ``jumps'' between $D$ possible discrete states throughout the trajectory. Marginal trajectories $p_t(x)$ are obtained by iterative sampling of position-wise transition distributions $p^i(x^i_{t+dt}|x_{t})$ that are categorical distributions parameterized by a learned sequence denoiser $p^{\theta}(x_1^i|x_t)$ trained to minimize a standard cross-entropy loss:

\begin{equation}
    \mathcal{L}_{CE} = \mathbb{E}_{t,p(z),p_t(x_t|z)} \left[ -\log p^{\theta}(x_1^i|x_t) \right]
\end{equation}

\section{Methods}

\subsection{Problem Setting} \label{sec:problem setting}

We represent a 3D molecule with $N$ atoms as a fully-connected graph. Each atom is a node in the graph. Every atom has a position in space $X = \{x_i\}_{i=1}^N \in \mathbb{R}^{N \times 3}$, an atom type (in this case the atomic element) $A = \{a_i\}_{i=1}^N $, and a formal charge $C = \{c_i\}_{i=1}^N $. Additionally, every pair of atoms has a bond order $E = \{  e_{ij} \forall i,j \in [N] | i \neq j \} $. Atom types, charges, and bond orders are categorical variables. When treated with continuous FM they are represented as one-hot encoded vectors. For brevity, we denote a molecule by the symbol $g$, which can be thought of as a tuple of the constituent data types $g = (X, A, C, E)$. 

There is no closed-form expression or analytical technique for sampling the distribution of realistic molecules $p(g)$. We seek to train a flow matching model to sample this distribution. We define the conditional probability path for a molecule to factorize into conditional probability paths over each modality. That is, conditional probability paths are defined independently for each modality.

\begin{equation}
    p_t(g|g_0,g_1)=p_t(X|X_0,X_1)p_t(A|A_0,A_1)p_t(C|C_0,C_1)p_t(E|E_0,E_1)
\end{equation}

We train one neural network to approximate $p(g_1|g_t)$ by jointly minimizing reconstruction losses for all data modalities. Our total loss is a weighted combination of FM losses on each modality:

\begin{equation}
    \mathcal{L} = \eta_X \mathcal{L}_{X} + \eta_A \mathcal{L}_{A} + \eta_C \mathcal{L}_{C} + \eta_E \mathcal{L}_{E}
\end{equation}

The raw output of the neural network is an estimate of the denoised molecule that will be obtained at the end of the trajectory. We denote the neural network outputs as $\hat{g}_1(g_t)$, and refer to this model as the ``denoiser''.

We train multiple variants of a \textit{de novo} molecule flow matching model using the same model architecture, datasets, and training procedure. Model variants are distinguished only by the discrete flow matching method used to generate atom types, atomic charges, and bond orders. 

\paragraph{DFM Variants} We train models with four different DFM methods. The ``Continuous'' model uses a fully continuous approach; making no accommodations for the discrete data. Categorical variables are generated via flows from a Gaussian prior to one-hot vectors; this is effectively the formulation of categorical flow matching presented in \citet{eijkelboom_variational_2024}. We implement continuous methods constrained to a continuous embedding of discrete data; these are Dirichlet Flows \cite{stark_dirichlet_2024} and SimplexFlow \cite{dunn_mixed_2024}. Finally we implement CTMC Flows \cite{campbell_generative_2024,gat_discrete_2024}, which are defined on discrete state spaces. Detailed descriptions of each flow matching method are presented in Appendix \ref{ap:fm-specifics}.

\subsection{Model Architecture}

We use the neural network architecture proposed by FlowMol \cite{dunn_mixed_2024}. Molecules are treated as fully-connected graphs. The model is designed to accept a sample $g_t$ and predict the final molecule $g_1$. FlowMol is comprised of stacks of sequential Molecule Update Blocks that update node features, node positions, and edge features. A single Molecule Update Block is composed of a message-passing graph convolution followed by node-wise and edge-wise updates. Geometric Vector Perceptrons (GVPs) \cite{jing_equivariant_2021} are used to learn and update vector features. The model architecture is explained in detail in Appendix~\ref{ap: arch}.

\subsection{Datasets and Model Evaluation}

We train models on GEOM-Drugs \cite{axelrod_geom_2022} using explicit hydrogens. GEOM-Drugs contains approximately 300k larger, drug-like molecules with multiple conformers for each molecule. We use the same dataset splits as \citet{vignac_midi_2023}. We also trained models on QM9 \cite{ruddigkeit_enumeration_2012,ramakrishnan_quantum_2014} but only present these results in the Appendix \ref{sec:qm9-results} because we consider it to be an overly simple benchmark of model performance. 

\paragraph{Validity} We report standard metrics on the validity of generated molecular topology: percent molecules stable and percent molecules valid. An atom is defined as ``stable'' if it has valid valency. Atomic valency is the sum of bond orders for an atom. A molecule is counted as stable if all of its constituent atoms are stable. A molecule is considered ``valid'' if it can be sanitized by RDKit \cite{noauthor_rdkit_nodate} using default sanitization settings.

\paragraph{Energy} Metrics regarding molecular topology provide no indication of quality of the molecular geometries produced by a model. Therefore, we also compute the Jensen-Shannon divergence of the distribution of potential energies for molecules in the training data and molecules sampled from models. Potential energies are obtained from the Merck Molecular Mechanics Force-Field implemented in RDKit \cite{noauthor_rdkit_nodate}. 

\paragraph{Functional Group Validity} Basic chemical validity is a necessary but insufficient condition for designing molecules for a particular application such as therapeutic use. Small-molecule drug discovery scientists have compiled sets of functional groups known to be unstable, toxic, or produce erroneous assay results. Taking inspiration from \citet{walters_generative_2024} we measure and report the presence of these problematic functional groups, which we call ``structural alerts,'' as a metric of molecule quality\footnote{Determining whether a functional group is problematic in the context of a drug discovery campaign can, in some cases, be subjective. However, measuring the presence of functional groups can still quantify similarity to training data at higher-order levels of organization than chemical valency.}. We specifically use the well-known Dundee \cite{brenk_lessons_2008} and Glaxo Wellcome \cite{hann_strategic_1999} structural alerts. Additionally, we count all of the unique ring systems observed in a batch of molecules. We then record how many times each unique ring system is observed in ChEMBL \cite{zdrazil_chembl_2024}, a database of 2.4M bio-active compounds. We report the average rate at which ring systems occur that are never observed in ChEMBL. Ring system and structural alert counting are implemented using the useful\_rdkit\_utils repository \cite{walters_patwaltersuseful_rdkit_utils_2024}. 

Molecule quality metrics are reported for samples of $10,000$ molecules, repeated 5 times. Inference is run on FlowMol using Euler integration with $100$ evenly-spaced time steps for QM9 and $250$ time steps for GEOM. All results are reported with $95\%$ confidence intervals. For all samplings, the number of atoms in each molecule is sampled according to the training data distribution.

\section{Experiments}

\paragraph{DFM ablations} The results of ablations in DFM type are shown in Table \ref{tab:geom-ablation}. The best performing method, by a large margin, is CTMC, producing an increase in molecular stability of 26 percentage points over the Continuous approach to discrete FM. Also notable in these results is that structural alerts and OOD rings are over-represented in generated molecules.
 
\begin{table}[t]
    \centering
    \caption{Discrete Flow Type Ablations on GEOM-Drugs}
    \label{tab:geom-ablation}
    \resizebox{\textwidth}{0.95\height}{
    \begin{tabular}{l|ccccc}
    \toprule
     \textbf{Categorical Flow Type}  & \thead{\textbf{Mols Stable} \\ \textbf{(\%) ($\uparrow$)}} & \thead{\textbf{Mols Valid} \\ \textbf{(\%) ($\uparrow$)}} & \thead{\textbf{JS(E)} \\ \textbf{($\downarrow$)}} & \thead{\textbf{Structural Alert Rate} \\ \textbf{(per mol) ($\downarrow$)}} & \thead{\textbf{OOD Ring } \\ \textbf{(per mol) ($\downarrow$)}} \\
    \midrule
    (Training Data) & $100$ & $100$ & $0$ & $0.69$ & $0.04$ \\
    \cline{1-6}
    Dirichlet & $20.4 { \scriptstyle \pm 0.3 }$ & $15.6 { \scriptstyle \pm 0.3 }$ & $0.45 { \scriptstyle \pm 0.01 }$ & $3.09 { \scriptstyle \pm 0.03 }$ & $0.64 { \scriptstyle \pm 0.01 }$ \\
    SimplexFlow & $64.6 { \scriptstyle \pm 0.3 }$ & $42.5 { \scriptstyle \pm 0.6 }$ & $0.33 { \scriptstyle \pm 0.01 }$ & $2.01 { \scriptstyle \pm 0.04 }$ & $0.32 { \scriptstyle \pm 0.01 }$ \\
    Continuous & $69.5 { \scriptstyle \pm 0.4 }$ & $51.3 { \scriptstyle \pm 0.5 }$ & $0.32 { \scriptstyle \pm 0.00 }$ & $2.39 { \scriptstyle \pm 0.03 }$ & $0.47 { \scriptstyle \pm 0.01 }$ \\
    CTMC & $ \mathbf{ 96.2 { \scriptstyle \pm 0.1 } }$ & $ \mathbf{ 91.5 { \scriptstyle \pm 0.3 } }$ & $ \mathbf{ 0.14 { \scriptstyle \pm 0.00 } }$ & $ \mathbf{ 1.23 { \scriptstyle \pm 0.01 } }$ & $ \mathbf{ 0.29 { \scriptstyle \pm 0.00 }}$ \\
    \cline{1-6}
    \bottomrule
    \end{tabular}
    } 

\end{table}

\paragraph{Explaining the Performance Gap} 

\begin{figure}[tb]
    \centering
    \includegraphics[width=0.95\textwidth]{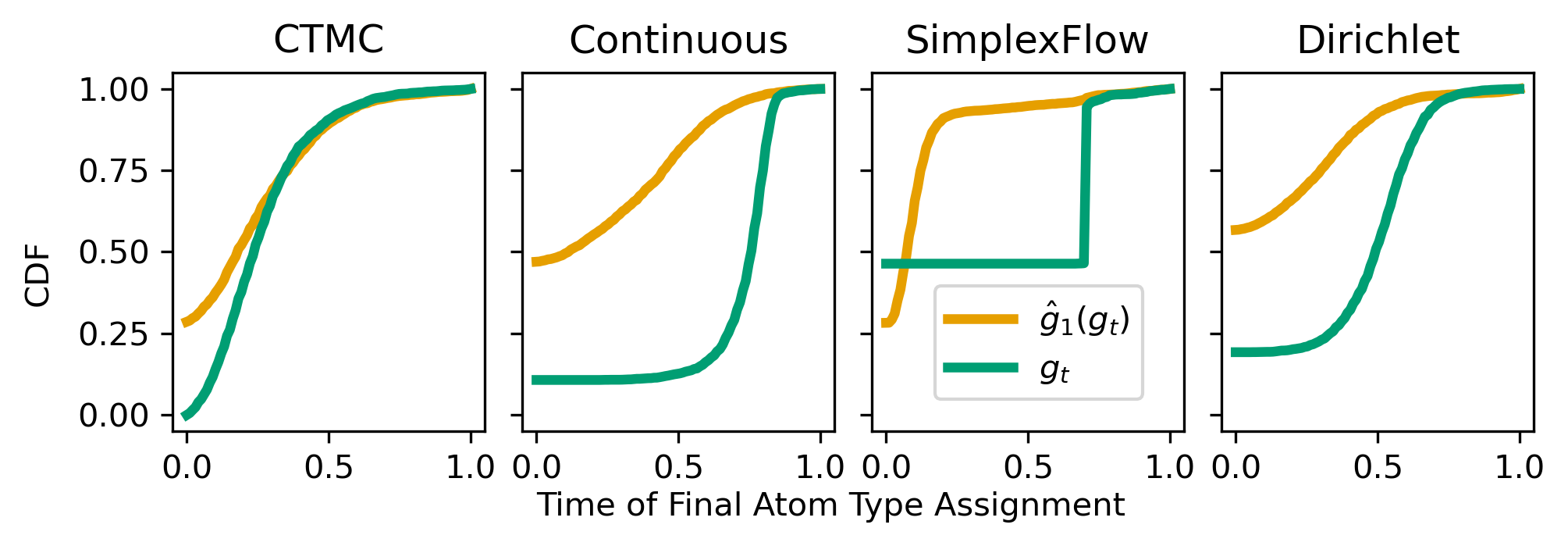}
    \caption{
    \textbf{Atom Type Assignment Times:} 
    Cumulative Density Functions (CDFs) of the time at which an atom is assigned its final atom type, for each DFM method tested. \textcolor{customgreen}{Green lines} show the time of final atom type assignments in $g_t$. \textcolor{customgold}{Gold lines} show the times when the final atom type is assigned in $\hat{g}_1(g_t)$ (the predicted final molecule given the current molecule at time $t$).
        \label{fig:switchdist}
    }
\end{figure}

\label{sec:perfgap} In the case of continuous flows on discrete data, a single atom's type vector moves towards a vertex of the simplex, e.g., a one-hot vector. The vertex that the atom type moves towards is the predicted atom type from the denoising model $\hat{g}_1(g_t)$ and the atom only gets close to the predicted vertex as $t \to 1$. As a result, there is a time lag between when the decoder $\hat{g}_1(g_t)$ makes an atom type assignment and when that assignment is reflected in $g_t$. We can quantify this phenomenon by measuring the time at which an atom acquires its final state in $g_t$ and $\hat{g}_1(g_t)$; we refer to this as the atom type assignment time. The distributions of assignment time in Figure \ref{fig:switchdist} show that continuous varieties of DFM exhibit substantial assignment time lagging. For example, at $t=0.5$ the Continuous DFM denoiser ($\hat{g}_1(g_t)$) has decided on the final atom type for 78\% of atoms while only 15\% of atoms in $g_t$ occupy their final atom type. State changes take a significant amount of time to occur when using continuous representations of discrete data. A similar pattern is seen in the assignment time distributions for SimplexFlow and Dirichlet flows as well. In contrast, because CTMC flows truly treat the data as discrete, atom types can often jump to the state predicted by the denoiser without delay; correspondingly, there is a smaller assignment time gap shown in Figure \ref{fig:switchdist}. CTMC flows are inherently better at making use of prior predictions and navigating the discrete spaces they model. Note that we use atom type only as an example of this pathology; this phenomenon occurs for any categorical data modeled with continuous flows. Additional details on this analysis are provided in Appendix \ref{ap:assignment_times}.

\paragraph{Comparison to Baselines}

We compare FlowMol-CTMC to three diffusion baselines that are considered SOTA for this task: MiDi \cite{vignac_midi_2023}, JODO \cite{huang_learning_2023}, and EQGAT-Diff \cite{le_navigating_2023}. Another relevant baseline is \citet{irwin_efficient_2024}; however, as no code or trained models are available independent evaluation is impossible and reported metrics are not directly comparable.

Performance of FlowMol-CTMC relative to baselines is shown in Table \ref{tab:baseline}. FlowMol-CTMC achieves SOTA performance on chemical valency metrics despite having fewer parameters. Interestingly, FlowMol-CTMC under performs baselines on functional group based metrics. All models express structural alerts at significantly higher rates than in the data. 

\begin{table}[h]
    \centering
    \caption{Comparison of FlowMol to baseline models on GEOM-Drugs}
    \label{tab:baseline}
    \resizebox{\textwidth}{0.95\height}{
    \begin{tabular}{l|cccccc}
         \toprule
     \textbf{Model}  & \thead{\textbf{Mols Stable} \\ \textbf{(\%) ($\uparrow$)}} & \thead{\textbf{Mols Valid} \\ \textbf{(\%) ($\uparrow$)}} & \thead{\textbf{JS(E)} \\ \textbf{($\downarrow$)}} & \thead{\textbf{Structural Alerts} \\ \textbf{(per mol) ($\downarrow$)}} & \thead{\textbf{OOD Rings} \\ \textbf{(per mol) ($\downarrow$)}} & \thead{\textbf{Parameters} \\ \textbf{($10^6$) ($\downarrow$)}} \\
         \midrule
        (Training Data) & $100$ & $100$ & $0$ & $0.69$ & $0.04$ & - \\
        \cline{1-7}
         MiDi & $ 85.1 { \scriptstyle \pm 0.9 }  $ & $ 71.6 { \scriptstyle \pm 0.9 }  $ & $ 0.23 { \scriptstyle \pm 0.00 }  $ & $ 1.01 { \scriptstyle \pm 0.01 }  $ & $ 0.33 { \scriptstyle \pm 0.00 }  $ & $24.1$ \\
         JODO & $ 90.7 { \scriptstyle \pm 0.5 }  $ & $ 76.5 { \scriptstyle \pm 0.8 }  $ & $ 0.17 { \scriptstyle \pm 0.01 }  $ & $ \mathbf{ 0.84 { \scriptstyle \pm 0.02 } } $ & $ \mathbf{0.21 { \scriptstyle \pm 0.00 } } $ & $5.7$ \\
         EQGAT-Diff &  $ 93.4 { \scriptstyle \pm 0.2 }  $ & $ 86.1 { \scriptstyle \pm 0.3 }  $ & $ \mathbf{ 0.11 { \scriptstyle \pm 0.00 } } $ & $ 1.06 { \scriptstyle \pm 0.01 }  $ & $ 0.27 { \scriptstyle \pm 0.00 }  $ & $12.3$ \\
         FlowMol-CTMC & $ \mathbf{ 96.2 { \scriptstyle \pm 0.1 } } $ & $ \mathbf{ 91.6 { \scriptstyle \pm 0.1 } } $ & $ 0.14 { \scriptstyle \pm 0.00 }  $ & $ 1.23 { \scriptstyle \pm 0.01 }  $ & $ 0.28 { \scriptstyle \pm 0.00 }  $ & $\mathbf{4.3}$ \\
         \bottomrule
    \end{tabular}
    } 

\end{table}

\section{Conclusions} \label{sec:conclusion}

We demonstrate that CTMC flows \cite{campbell_generative_2024,gat_discrete_2024} are the most effective discrete flow matching method evaluated; changing only the DFM formulation has dramatic effects on the quality of generated samples. We also present a compelling mechanistic explanation for the performance gap between CTMCs and continuous approaches to discrete flow matching. Our analysis reinforces the idea that, generally, relaxing discrete data to continuous spaces introduces undesirable pathologies. As a result of this analysis, we present FlowMol-CTMC, a 3D de novo molecule generative model achieving state of the art molecular validity with fewer learnable parameters than comparable methods. 

Finally, we introduce novel metrics for the quality of de novo designed molecules by measuring the presence of problematic functional groups and unusual ring systems. Our comparison to previous state of the art models shows that a model can have improved validity while producing more problematic functional groups; making clear that future work in this field should move beyond validity-based metrics and towards higher-order notions of molecular quality. 

\section{Acknowledgments}

This work is funded through R35GM140753 from the National Institute of General Medical Sciences. The content is solely the responsibility of the authors and does not necessarily represent the official views of the National Institute of General Medical Sciences or the National Institutes of Health.

\typeout{} 
\bibliography{main}

\appendix

\section{Model Formulation} \label{ap:fm-specifics}

\subsection{Atomic Coordinate Flows}

The conditioning variable for atomic coordinate flows is paired initial and final atomic positions $z=(X_0,X_1)$. The distribution of the conditioning variable, often referred to as the coupling distribution \cite{tong_improving_2023, gat_discrete_2024} is the equivariant optimal transport coupling $(X_0,X_1) \sim \pi^*(X_0,X_1)$ \cite{klein_equivariant_2023,song_equivariant_2023}. This means that $X_0$ and $X_1$ are independently sampled from the prior and data distribution, respectively, before being aligned via rigid body rotation, translation, and a distance minimizing permutation of node assignments. The conditional probability path for positions is a Dirac density placed on a straight line connecting the terminal states.

\begin{equation} \label{eq:xcondpath}
    p_t(X|X_0,X_1) = \delta(X - (1 - \kappa_t)X_0 - \kappa_t X_1)
\end{equation}

This is equivalent to a deterministic interpolant \cite{albergo_stochastic_2023}:

\begin{equation} \label{eq:x-interpolant}
    X_t = (1 - \kappa_t)X_0 + \kappa_t X_1
\end{equation}

 Where $\kappa_t: [0,1] \to [0,1]$ is the ``interpolant schedule'': a function taking $t$ as input and returning a number between 0 and 1. The interpolant schedule controls the rate at which atomic coordinates transition from the prior to the target distribution. We design the conditional paths of each modality to have an interpolant schedule, and these interpolant schedules need not be the same for each modality. Interpolant schedule choices are discussed in Appendix \ref{ap:interpsched}.
 
 The prior distribution for atomic coordinates are i.i.d. samples from a standard Gaussian $p_0(X)=\prod_{i=1}^N \mathcal{N}(x^i_0| 0, \mathbb{I})$. 

 The conditional probability path \eqref{eq:x-interpolant} is produced by the conditional vector field \eqref{eq:continuous-cond-vec-field} \cite{dunn_mixed_2024,gat_discrete_2024,tong_improving_2023}.

 \begin{equation} \label{eq:continuous-cond-vec-field}
     u_t(X|X_0,X_1) = \frac{\dot{\kappa}_t}{1-\kappa_t}\left( X_1 - X_t \right)
 \end{equation}

 Because of this, it can be shown that the marginal distribution $p_t(X)$ can be sampled by numerical integration of the vector field \eqref{eq:learnvecfield} \cite{dunn_mixed_2024}

 \begin{equation} \label{eq:learnvecfield}
     u(X_t) = \frac{\dot{\kappa}_t}{1-\kappa_t}\left( \hat{X}_1(X_t) - X_t \right)
 \end{equation}

 Where $\dot{\kappa}_t$ is the time derivative of the interpolant schedule. $\hat{X}_1(X_t)$ is a neural network trained under the denoising loss \eqref{eq:denoiseloss}.

 \begin{equation} \label{eq:denoiseloss}
    \mathcal{L} = \mathbb{E}_{t,p_t(X_t|X_0,X_1),\pi^*(X_0,X_1)} 
    \left[ 
    \norm{\hat{X}_1(X_t) - X_t}
    \right]
 \end{equation}

 In practice $\hat{X}_1$ predictions are obtained as an output head of a neural network that takes in a partially formed molecule $g_t$ and simultaneously predicts $(\hat{X}_1,\hat{A}_1,\hat{C}_1,\hat{E}_1) = \hat{g}_1(g_t)$.

\subsection{Continuous Flows on Discrete Data} \label{ap:condescflow}

For the ``Continuous'' model variant shown in Table \ref{tab:geom-ablation}, categorical variables are treated as fully continuous. We describe here the formulation using atom types as an example but the equations apply equivalently to atom charges and bond orders.

The conditioning variable for fully continuous flows is pairs of initial and terminal states $z=(A_0,A_1)$. The coupling distribution $(A_0,A_1) \sim \pi(A_0,A_1)$ is the independent coupling $\pi(A_0,A_1) = p_0(A_0)p_1(A_1)$; samples are drawn independently from the prior and data distribution.

Each atom type is considered a vector of real numbers $a^i_t \in \mathbb{R}^{n_a}$ where $n_a$ is the number of atom types supported. The prior distribution for atom types are IID samples from a standard Gaussian: $p(A_0)=\prod_{i=1}^N p(a^i_0)=\prod_{i=1}^N \mathcal{N}(a_0^i|0, \mathbb{I})$. The data distribution are one-hot vectors indicating the type of each atom. 

The conditional probability path is the same used for atomic coordinate flows \eqref{eq:continuous-cond-vec-field}. The learned vector field for sampling $p_t(A_t)$ is of the same form as well \eqref{eq:learnvecfield}. However, we use a softmax activation layer guaranteeing predicted endpoints are on the simplex. Correspondingly the denoising loss is a cross-entropy loss rather than an MSE loss. 

\subsection{Continuous Flows on a Continuous Embedding of Discrete Data} \label{ap:contflow}

To design flow matching for categorical data, the strategy of SimplexFlow \cite{dunn_mixed_2024} and Dirichlet Flows \cite{stark_dirichlet_2024} is to define a continuous representation of categorical variables, and then construct a flow matching model where flows are constrained to this representation. The continuous representation chosen for a $d$-categorical variable is the d-dimensional probability simplex $\mathcal{S}^d$:

\begin{equation}
    \mathcal{S}^d = \left\{ x \in \mathbb{R}^d | x_i \geq 0, \id \cdot x=1  \right\}
\end{equation}

A $d$-categorical variable $x_1 \in \{ 1,2,\dots,d \}$ can be converted to a point on $\mathcal{S}^d$ via one-hot encoding. Correspondingly, the categorical distribution $p_1(x) = \mathcal{C}(q)$ can be converted to a distribution on $\mathcal{S}^d$:

\begin{equation} \label{eq:simplex-data-dist}
    p_1(x) = \sum_{j=1}^d q_i\delta(x - e_j)
\end{equation}

where $e_j$ is the $j^{th}$ vertex of the simplex and $q_j$ is the probability of $x$ belonging to the $j^{th}$ category.

\subsubsection{SimplexFlow}

SimplexFlow \cite{dunn_mixed_2024} uses the same conditional probability path \eqref{eq:xcondpath} and learned vector field \eqref{eq:learnvecfield} as for atomic coordinate flows. The training objective for categorical feature denoising is the cross entropy loss as described in Section \ref{ap:contflow}. 

\begin{figure}[h]
    \centering
    \includegraphics[width=0.75\textwidth]{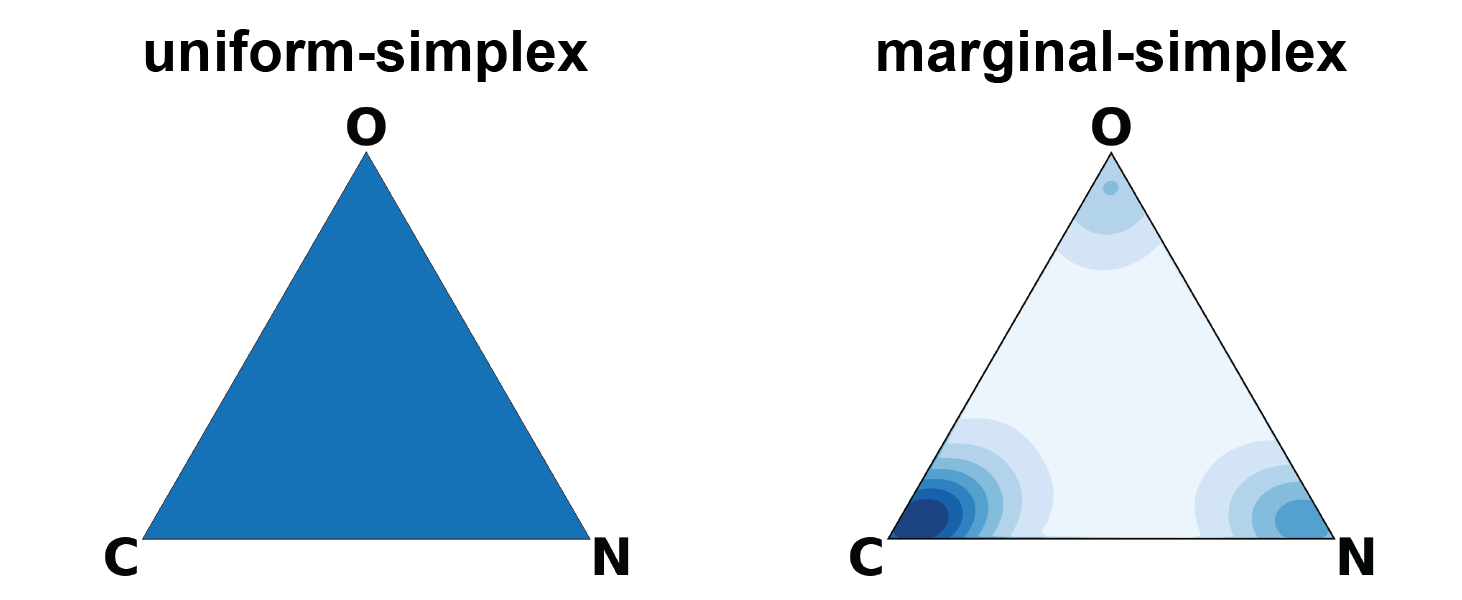}
    \caption{\label{fig:simprior}
    Prior distributions used with support on the probability simplex
    }
\end{figure}

The prior distribution used is the ``marginal-simplex'' prior. That is, atom types are first sampled as one-hot vectors according to their marginal distribution in the training data. We then apply a blurring procedure by Gaussian noise addition and projection back on to the simplex. The marginal-simplex prior is visualized in Figure \ref{fig:simprior}

\subsubsection{Dirichlet Flows}

For Dirichlet Flows \cite{stark_dirichlet_2024} the conditional probability path for a single atom $a^i_t$ is 

\begin{equation}
    p_t(a^i|a^i_1=e^j) = \mathrm{Dir}(a^i|\gamma = 1 + e^j\omega)
\end{equation}

Where $\mathrm{Dir}$ is a Dirichlet distribution parameterized by $\gamma$ and $\omega$ represents time. The Dirichlet conditional flow must start at $\omega = 1$ and only converges to $\delta(a^i-e^j)$ in the limit $\omega \to \infty$. In order to incorporate Dirichlet flows into our model, we define the relation $\omega_t = \omega_{\mathrm{max}}\kappa_t + 1$, where $\kappa_t$ is the interpolant schedule. We set $\omega_{\mathrm{max}} = 10$. Dirichlet flow matching necessitates the use of a uniform prior over the simplex which is visualized in Figure \ref{fig:simprior}. 

The training objective is a cross-entropy loss on category logits. We refer the reader to \cite{stark_dirichlet_2024} for a description of the generating conditional vector field. The marginal vector field is computed as an expectation over conditional vector fields with respect to $p^\theta(a^i_1|a^i_t)$.

\subsection{CTMC Flows} 

When we model discrete data with CTMC flows \cite{campbell_generative_2024, gat_discrete_2024}, we introduce an additional category, the ``mask'' token. Our molecule atom types are considered as a sequence of discrete tokens $A = (a^1, a^2, \dots, a^N)$ where $N$ is the number of atoms in the molecule and $a^i \in \{1,2,\dots,n_a,M\}$ where $n_a$ is the number of supported atom types and $M$ is the mask token. The prior distribution is Kronecker delta placed on a sequence of mask states, i.e., $p_0(A) = \prod_{i=1}^N \delta_M(a^i)$. Conditional probability paths factorize over each atom in the molecule:

\begin{equation}
    p_t(A|A_0,A_1) = \prod_{i=1}^N p_t^i(a^i|A_0,A_1) 
\end{equation}

Where the per-atom conditional probability path takes the form:

\begin{equation} \label{eq:ctmc-condpath}
    p_t^i(a^i|A_0,A_1) = \kappa_t\delta_{A_1}(a^i) + (1-\kappa_t)\delta_{M}(a^i)
\end{equation}

Where $M$ is the mask token. In other words, at time $t$, atom $a^i$ has probability $\kappa_t$ of being in the masked state and probability $1 - \kappa_t$ of being in its final state $a^i_1$. Note that we also use an abbreviated notation here where $\delta_{A_1}(a^i) = \delta_{a_1^i}(a^i) $.

Atom type trajectories are sampled by iterative sampling of transition distributions that factorize over atoms:

\begin{equation}
    p(A_{t+\Delta t}|A_t) = \prod_{i=1}^N p^i(a^i_{t+\Delta t}|A_t)
\end{equation}

The per-atom transition distributions are categorical distributions with logits:

\begin{equation} \label{eq:ctmctrans}
    p^i(a^i_{t+\Delta t}=j|A_t) = \delta_{a^i_t}(j) + u^i(j,A_t)\Delta t
\end{equation}

Where $u^i(j,A_t) \in \mathbb{R}$ is the instantaneous flow of probability towards atom type $j$ for atom $i$. \citet{campbell_generative_2024} refers to this quantity as a rate matrix and \citet{gat_discrete_2024} calls it the probability velocity. 

The primary difference between the CTMC formulations in \citet{campbell_generative_2024} and \citet{gat_discrete_2024} are the form of the probability velocity used to generate trajectories. For our application, we find the probability velocity proposed by \citet{campbell_generative_2024} to give better results; this is contradictory to the conclusions of \citet{gat_discrete_2024}.

Using the proposed conditional rate matrix of \citet{campbell_generative_2024}, our conditional probability path \eqref{eq:ctmc-condpath} induces the following marginal probability velocity for $j \neq a^i_t$:

\begin{equation}
    u^i(j,A_t) = \frac{\dot{\kappa}_t + \eta \kappa_t}{1 - \kappa_t} p_{1|t}^{\theta}(a_1^i = j^i | A_t) \delta_M(a_t^i) + \eta (1 - \delta_M(a_t^i)) \delta_M(j)
\end{equation}

Where $\eta$ is the stochasticity parameter that, when increased, increases both the rate at which tokens are unmasked and remasked. To ensure that \eqref{eq:ctmctrans} defines a valid categorical distribution we set $u^i(a_t^i,A_t)= -\sum_{j\neq a_i^t}u^i(j,A_t)$.

Plugging $u^i(a_t^i,A_t)$ into the transition distribution \eqref{eq:ctmctrans} yields sampling dynamics that can be described very simply. If $a_i^t = M$, the probability of unmasking is $\Delta t\frac{\dot{\kappa}_t + \eta \kappa_t}{1 - \kappa_t}$. If we do unmask, the unmasked state is selected according to our learned model $p_{1|t}^{\theta}(a_1^i | A_t)$. If we are currently in an unmasked state, the probability of switching to a masked state is $\eta \Delta t$.

We find low-temperature sampling, where the prediction logits are re-normalized with temperature $\tau$:

\begin{equation}
    p_{1|t}^{\theta}(a_1^i | A_t) = \text{softmax}\left(\tau^{-1} \log{p_{1|t}^{\theta}(a_1^i | A_t)} \right)
\end{equation}

to be critical for model performance. In practice we use $\tau = 0.05$ and $\eta = 30$.

\subsection{Interpolant Schedules} \label{ap:interpsched}

When using CTMC flows for categorical data, we find a linear interpolant $\kappa_t = t$ for all modalities gives strong performance. For all other flows, we find that a cosine interpolant works better:

\begin{equation} \label{eq:cos-sched}
    \kappa_t = 1 - \cos^2 \left(\frac{\pi}{2}t^{\nu}
    \right)
\end{equation}

where separate values of $\nu$ are used for each modality. We use the same values as described in \cite{dunn_mixed_2024}. Using separate interpolants for each modality encourages the model to determine positions before atom types, before bond orders, and so on. The reason this specifically benefits continuous FM approaches is likely due to the pathologies outlined in Section \ref{sec:perfgap}.

\section{Architecture} \label{ap: arch}

\begin{figure}[H]
    \centering
    \includegraphics[width=0.8\textwidth]{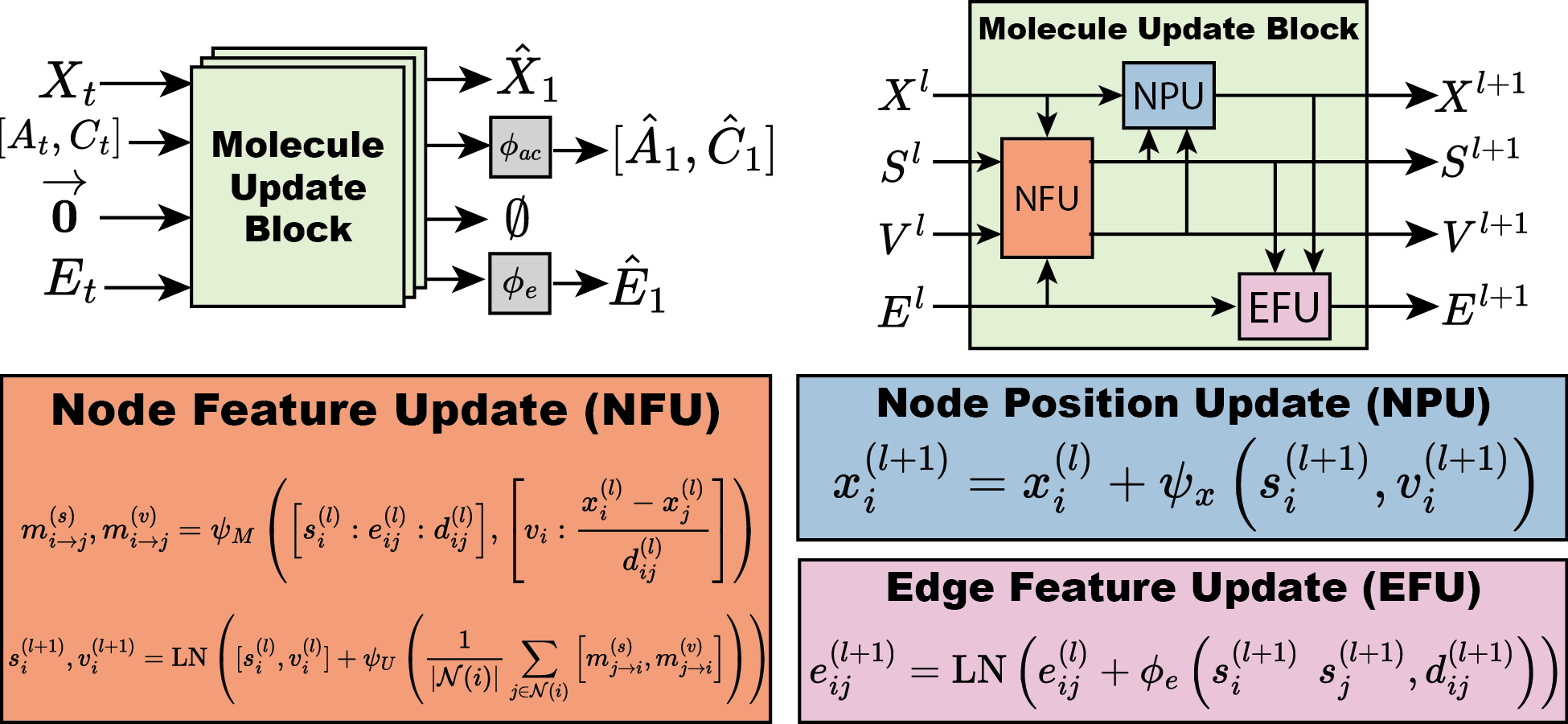}
    \caption{\textbf{FlowMol Architecture} \textit{Top left}: An input molecular graph $g_t$ is transformed into a predicted final molecular graph $g_1$ by being passed through multiple molecule update blocks. \textit{Top right:} A molecule update block uses NFU, NPU, and EFU sub-components to update all molecular features. \textit{Bottom:} Update equations for graph features. $\phi$ and $\psi$ denote MLPs and GVPs, respectively.  }
    \label{fig:arch}
\end{figure}

\paragraph{Overview} Molecules are treated as fully-connected graphs. The model is designed to accept a sample $g_t$ and predict the final destination molecule $g_1$. Within the neural network, molecular features are grouped into node positions, node scalar features, node vector features, and edge features. Node positions are identical to the atom positions discussed in Section \ref{sec:problem setting}. Node scalar features are a concatenation of atom type and atom charge. Node vector features are geometric vectors (vectors with rotation order 1) that are relative to the node position. Node vector features are initialized to zero vectors. Molecular features are iteratively updated by passing $g_t$ through several Molecule Update Blocks. A Molecule Update Block uses Geometric Vector Perceptrons (GVPs) \cite{jing_equivariant_2021} to handle vector features. Molecule Update Blocks are composed of three components: a node feature update (NFU), node position update (NPU) and edge feature update (EFU). The NFU uses a message-passing graph convolution to update node features. The NPU and EFU blocks are node and edge-wise operations, respectively. Following several molecule update blocks, predictions of the final categorical features ($\hat{A}_1, \hat{C}_1, \hat{E}_1$) are generated by passing node and edge features through shallow node-wise and edge-wise multi layer perceptrons (MLPs). For models using endpoint parameterization, these MLPs include softmax activations. The model architecture is visualized in Figure \ref{fig:arch}. 

FlowMol is implemented using PyTorch and the Deep Graph Library (DGL) \cite{wang_deep_2020}.

Each node is endowed with a position in space $x_i \in \mathbb{R}^3$, scalar features $s_i \in \mathbb{R}^d$, and vector features $v_i \in \mathbb{R}^{c \times 3}$. Scalar features are initialized at the network input by concatenating atom type and charge vectors: $s_i^{(0)} = [a_i:c_i]$. Vector features are initialized to zeros $v_i^{(0)} = \mathbf{0}$.
Each edge is endowed with scalar edge features that, at the input to the network, are the bond order at time $t$. We enforce that the bond order on both edges for a pair of atoms is identical: $e_{ij}^{(0)} = e_{ji}^{(0)}$.

In practice, graphs are directed. For every pair of atoms $i,j$ there exists edges in both directions: $i \rightarrow j$ and $j \rightarrow i$. When predicting the final bond orders $\hat{E}_1$ for an edge, we ensure that one prediction is made per pair of atoms and that this prediction is invariant to permutations of the atom indexing. This is accomplished by making our prediction from the sum of the learned bond features. That is, $\hat{e}_1^{ij} = MLP( e_{ij} + e_{ji} )$. 

\paragraph{Molecule Update Block} We define a Molecule Update Block that will update all graph features $x_i, s_i, v_i, e_{ij}$. Each molecule update block is comprised of 3-sub blocks: a node feature update block, a node position update block, and an edge feature update block. The input molecule graph is passed through $L$ Molecule Update Blocks. Vector features are operated on by geometric vector perceptions (GVPs). A detailed description of our implementation of GVP is provided in Section \ref{sec:gvp}.

\paragraph{Node Feature Update Block} The node feature update block will perform a graph convolution to update node scalar and vector features $s_i, v_i$. The message generating and node-update functions for this graph convolution are each chains of GVPs. GVPs accept and return a tuple of scalar and vector features. Therefore, scalar and vector messages $m_{i \to j}^{(s)}$ and $m_{i \to j}^{(v)}$ are generated by a single function $\psi_M$ which is two GVPs chained together.

\begin{equation}
    \label{gvp_mij}
    m_{i \to j}^{(s)}, m_{i \to j}^{(v)} = \psi_M \left(
    \left[ s_i^{(l)} : e_{ij}^{(l)} : d_{ij}^{(l)} \right]
    , \left[v_i : \frac{x_i^{(l)} - x_j^{(l)}}{d_{ij}^{(l)}} \right]\right) 
\end{equation}

Where $:$ denotes concatenation, and $d_{ij}$ is the distance between nodes $i$ and $j$ at molecule update block $l$. In practice, we replace all instances of $d_{ij}$ with a radial basis embedding of that distance before passing through GVPs or MLPs. Message aggregation and node features updates are performed as described in \cite{jing_equivariant_2021}, with the exception that we do not use a dropout layer:

\begin{equation}
    \label{gvp_update}
    s_i^{(l+1)}, v_i^{(l+1)} = \mathrm{LayerNorm} \left( [s_i^{(l)}, v_i^{(l)}] + \\
     \psi_U \left( \frac{1}{|\mathcal{N}(i)|} \sum_{j \in \mathcal{N}(i) } \left[ m_{j \to i}^{(s)}, m_{j \to i}^{(v)} \right]   \right) \right) 
\end{equation}

The node update function $\psi_U$ is a chain of three GVPs.

\paragraph{Node Position Update Block} The purpose of this block is to update node positions $x_i$. Node positions are updated as follows:

\begin{equation}
    x_i^{(l+1)} = x_i^{(l)} + \psi_x \left( s_i^{(l+1)}, v_i^{(l+1)} \right)
\end{equation}

Where $P$ is a chain of 3 GVPs in which the final GVP emits 1 vector and 0 scalar features. Moreover for the final GVP, the vector-gating activation function ($\sigma_g$ in Algorithm \ref{alg:gvp}), which is typically a sigmoid function, is replaced with the identity.

\paragraph{Edge Feature Update Block} Edge features are updated by the following equation:

\begin{equation}
       e_{ij}^{(l+1)} = \mathrm{LayerNorm} \left(e_{ij}^{(l)} + \phi_e \left( s_i^{(l+1)}, s_j^{(l+1)}, d_{ij}^{(l+1)} \right) \right)
\end{equation}

Where $\phi_e$ is a shallow MLP that accepts as input the node scalar features of nodes participating in the edge as well as the distance between the nodes from the positions compute in the NPU block.

\subsection{GVP with Cross Product} \label{sec:gvp}

A geometric vector perception (GVP) can be thought of as a single-layer neural network that applies linear and point-wise non-linear transformation to its inputs. The difference between a GVP and a conventional feed-forward neural network is that GVPs operate on two distinct data types: scalars and vectors. GVPs also allow these data types to exchange information while preserving equivariance of the output vectors. The original GVP only applied linear transformations to the vector features and as a result produces output vectors that are E(3)-equivariant. 

We introduce a modification to the GVP as its presented in \citet{jing_equivariant_2021}; specifically we perform a cross product operation on the input vectors. The motivation for this is that the cross product is \textit{not} equivariant to reflections. As a result, the version of GVP we present here is SE(3) equivariant. The benefit of being SE(3) equivariant rather than E(3) equivariant is that the model becomes sensitive to chiral centers in molecules, since reflecting the molecule will produce different model outputs. The operations for our cross product enhanced GVP are described in Algorithm \ref{alg:gvp}.

\begin{algorithm}[t]
\caption{Geometric Vector Perceptron with Cross Product} 
\label{alg:gvp}
\begin{algorithmic}
\setstretch{1.2}
\STATE
\STATE \textbf{Input:} Scalar and vector features: $(s,v) \in \mathbb{R}^f \times \mathbb{R}^{\nu \times 3}$
\STATE \textbf{Output:} Scalar and vector features: $(s',v') \in \mathbb{R}^j \times \mathbb{R}^{\mu \times 3}$
\STATE \textbf{Hyperparameter:} Number of hidden vector features $n_h \in \mathbb{Z}^{+}$ 
\STATE \textbf{Hyperparameter:} Number of cross product features $n_{cp} \in \mathbb{Z}^{+}$ 
\STATE \ \ \ \ $v_{h} \gets W_h v \quad \in \mathbb{R}^{n_h \times 3} $
\STATE \ \ \ \ $v_{cp} \gets W_{cp} v \quad \in\mathbb{R}^{2n_{cp} \times 3}$
\STATE \ \ \ \ $v_{cp} \gets v_{cp}[:n_{cp}] \times v_{cp}[n_{cp}:] \quad \in \mathbb{R}^{n_{cp} \times 3}  $ // cross product
\STATE \ \ \ \ $ v_{h+cp} \gets \mathrm{Concat}(v_h, v_{cp}) \quad \in \mathbb{R}^{(n_h + n_{cp}) \times 3} $ // concatenation along rows
\STATE \ \ \ \ $v_{\mu} \gets W_{\mu} v_{h+cp} \quad \in \mathbb{R}^{\mu \times 3} $
\STATE \ \ \ \ $ s_{h+cp} \gets \norm{v_{h+cp}} \quad \in\mathbb{R}^{n_h + n_{cp}} $ 
\STATE \ \ \ \ $ s_{f+h+cp} \gets \mathrm{Concat}(s, s_{h+cp})$'
\STATE \ \ \ \ $s_j \gets W_j s_{f+h+cp} + b_j \quad \in \mathbb{R}^j$
\STATE \ \ \ \ $s' \gets \sigma(s_j) \quad \in \mathbb{R}^{j}$
\STATE \ \ \ \ $v' \gets { \sigma_g\left(W_g[\sigma^+(s_m)]+b_g\right)\odot v_\mu} \ (\text{row-wise}) \quad \in \mathbb{R}^{\mu \times 3}$ 
\STATE  \textbf{return} $(s', v')$  
\end{algorithmic}
\end{algorithm}

\section{Atom Type Assignment Time Analysis} \label{ap:assignment_times}

\begin{figure}[t]
    \centering
    
    \includegraphics[width=0.95\textwidth]{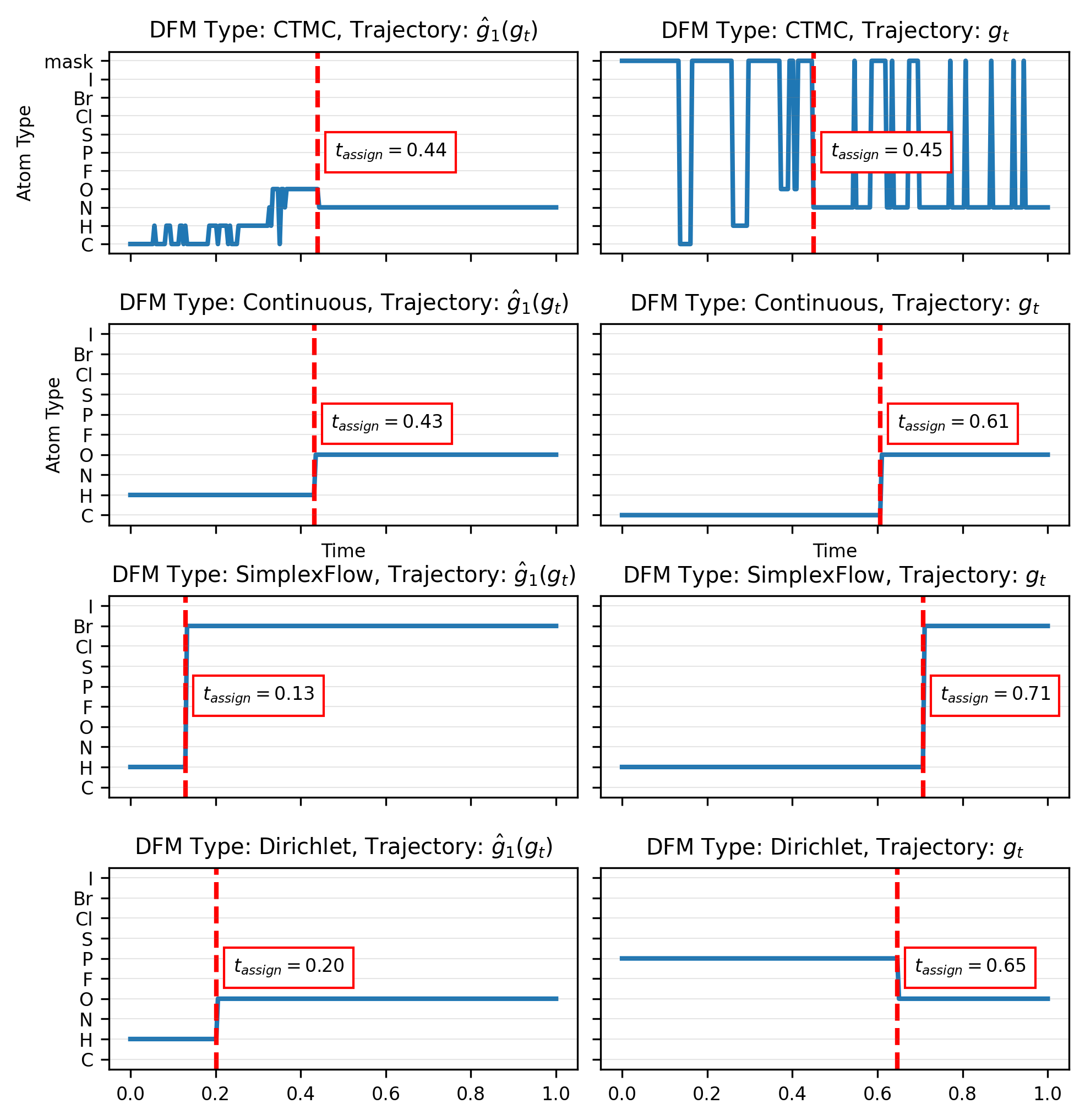}
    \caption{ \label{fig:atom_traj_example}
    Atom type trajectories and associated times of final type assignment. Each plot contains the atom type trajectory for a single atom. Subplot columns correspond to the trajectory type: either a $\hat{g}_1(g_t)$ or $g_t$ trajectory. Subplot rows correspond to the DFM variant used by the model. Within a row, the $\hat{g}_1(g_t)$ and $g_t$ trajectories displayed are for the same atom. The time of final atom type assignment is overlaid in red on each trajectory. 
    }
\end{figure}

For a sampled molecule, we obtain two trajectories: the state of the molecule at every time step $g_t$ and the output of the denoiser at every time step $\hat{g}_1(g_t)$. For both of the these trajectories, we record the time of the final type assignment for each atom; this is the time at which an atom occupies its final state and remains in that final state for the remainder of the trajectory. For clarity, Figure \ref{fig:atom_traj_example} shows atom type trajectories for single atoms and shows where the time of final assignment, $t_{assign}$, occurs in each trajectory.

\vspace{-1pt}

For all DFM variants other than CTMCs, atom type vectors are not necessarily one-hot vectors during a trajectory. For the analyses presented in Figure \ref{fig:switchdist} and Figure \ref{fig:atom_traj_example}, atom type vectors are assigned to discrete states by finding the one-hot vector that is closest, as determined by Euclidean distance, to the current atom type vector. 

When computing atom type assignment times for CTMC flows in $g_t$, masked states are not counted as state changes contributing to $t_{assign}$; an atom is only considered to switch states when it enters an unmasked state that is different from its previous masked state. 

To generate Figure \ref{fig:switchdist}, we sampled 100 molecules from each of the four DFM variants of FlowMol. The number of atoms in each molecule was sampled from the atom count distribution from the training data and kept the same across models.

\section{Model Details}

QM9 models are trained with 8 Molecule Update Blocks while GEOM models are trained with 5. Atoms contain 256 hidden scalar features and 16 hidden vector features. Edges contain 128 hidden features. QM9 models are trained for 1000 epochs and GEOM models are trained for 20 epochs. QM9 models are trained on a single L40 GPU with a batch size of 128. GEOM models are trained on 4xL40 GPUs with a per-GPU batch size of 16. QM9 models train in about 3 days while GEOM models take 4-5 days. All model hyperparameters are visible in the config files provided in our github repository. 

$(\eta_{X}, \eta_{A}, \eta_{C}, \eta_{E})$ are scalars weighting the relative contribution of each loss term. We set these values to $(3, 0.4, 1, 2)$ as was done in \citet{vignac_midi_2023}. 

\section{QM9 results} \label{sec:qm9-results}

The QM9 dataset \cite{ruddigkeit_enumeration_2012,ramakrishnan_quantum_2014} contains 124k small molecules, each with one 3D conformation. Molecules in QM9 have an average of 18 atoms and a max of 29. Results for DFM ablations on QM9 are shown in Table \ref{tab:qm9-ablation}. FlowMol-CTMC performance relative to baselines on QM9 is shown in Table \ref{tab:qm9baseline}

\begin{table}[H]
    \centering
    \caption{Discrete Flow Type Ablations on QM9}
    \label{tab:qm9-ablation}
    \begin{tabular}{l|ccc}
    \toprule
     \textbf{Discrete Flow Type}  & \thead{\textbf{Mols Stable} \\ \textbf{(\%) ($\uparrow$)}} & \thead{\textbf{Mols Valid} \\ \textbf{(\%) ($\uparrow$)}} & \thead{\textbf{JS(E)} \\ \textbf{($\downarrow$)}} \\
    \midrule
    Dirichlet & $85.0 { \scriptstyle \pm 1.0 }$ & $87.4 { \scriptstyle \pm 0.5 }$ & $0.12 { \scriptstyle \pm 0.01 }$ \\
    SimplexFlow & $87.9 { \scriptstyle \pm 0.3 }$ & $92.3 { \scriptstyle \pm 0.2 }$ & $\mathbf{0.07 { \scriptstyle \pm 0.00 }}$ \\
    Continuous & $96.8 { \scriptstyle \pm 0.4 }$ & $96.2 { \scriptstyle \pm 0.3 }$ & $0.10 { \scriptstyle \pm 0.01 }$ \\
    CTMC & $ \mathbf{ 99.3 { \scriptstyle \pm 0.1 }}$ & $\mathbf{99.3 { \scriptstyle \pm 0.1 }}$ & $0.09 { \scriptstyle \pm 0.00 }$ \\
    \cline{1-4}
    \bottomrule
    \end{tabular}

\end{table}

\begin{table}[H]
    \centering
    \caption{Comparison of FlowMol to baseline models on QM9}
    \label{tab:qm9baseline}
    \begin{tabular}{l|ccc}
         \toprule
     \textbf{Model}  & \thead{\textbf{Mols Stable} \\ \textbf{(\%) ($\uparrow$)}} & \thead{\textbf{Mols Valid} \\ \textbf{(\%) ($\uparrow$)}} & \thead{\textbf{JS(E)} \\ \textbf{($\downarrow$)}} \\
         MiDi & $ 97.5 { \scriptstyle \pm 0.1 }  $ & $ 98.0 { \scriptstyle \pm 0.2 }  $ & $ \mathbf{ 0.05 { \scriptstyle \pm 0.00 } } $ \\
         JODO & $ 98.7 { \scriptstyle \pm 0.2 }  $ & $ 98.9 { \scriptstyle \pm 0.2 }  $ & $ 0.12 { \scriptstyle \pm 0.01 }  $ \\
         EQGAT-Diff &  $ 98.8 { \scriptstyle \pm 0.1 }  $ & $ 99.0 { \scriptstyle \pm 0.0 }  $ & $ 0.08 { \scriptstyle \pm 0.00 }  $ \\
         FlowMol-CTMC & $ \mathbf{ 99.3 { \scriptstyle \pm 0.1 } } $ & $ \mathbf{ 99.3 { \scriptstyle \pm 0.1 } } $ & $ 0.10 { \scriptstyle \pm 0.00 }  $ \\
         \bottomrule
    \end{tabular}

\end{table}

\end{document}